%% file: root.tex
\documentclass[letterpaper, 10 pt, conference]{ieeeconf}  
\IEEEoverridecommandlockouts                              
\input{sections/package}

\title{\LARGE \bf
Optimized Design of A Haptic Unit for Vibrotactile Amplitude Modulation
}

\author{Jingchen Huang, Yun Fang, Weichao Guo, and Xinjun Sheng$^{\dag}$            
\thanks{ This work is supported in part by the National Natural Science Foundation of China (Grant Nos. 52175021, 52375021), in part by State Key Laboratory of Robotics and Systems (HIT) (Grant No. SKLRS-2024-KF-01). The authors are with the State Key Laboratory of Mechanical System and Vibration, School of Mechanical Engineering, and the Meta Robotics Institute, Shanghai Jiao Tong University, Shanghai 200240, China (e-mail: huang\_jc@sjtu.edu.cn; fangyun@sjtu.edu.cn; guoweichao90@gmail.com; xjsheng@sjtu.edu.cn)}
\thanks{$^{\dag}$Corresponding author: Xinjun Sheng}
}

\begin{document}
\maketitle
\thispagestyle{empty}
\pagestyle{empty}

\begin{abstract}
Communicating information to users is a crucial aspect of human-machine interaction. Vibrotactile feedback encodes information into spatiotemporal vibrations, enabling users to perceive tactile sensations. It offers advantages such as lightweight, wearability, and high stability, with broad applications in sensory substitution, virtual reality, education, and healthcare. However, existing haptic unit designs lack amplitude modulation capabilities, which limits their applications. This paper proposed an optimized design of the haptic unit from the perspective of vibration amplitude modulation. A modified elastic model was developed to describe the propagation and attenuation mechanisms of vibration in the skin. Based on the model, two types of hierarchical architectural design were proposed. The design incorporated various materials arranged in multiple layers to amplify or attenuate the vibration amplitude as it traveled through the structure. An experimental platform was built to evaluate the performance of the optimized design.
\end{abstract}


\section{INTRODUCTION}
When robots are operated to complete tasks, it is crucial not only to understand users’ intentions accurately but also to provide feedback through various sensory systems, including visual, auditory, and tactile feedback. Among these, tactile feedback, which stimulates the skin—the largest organ of the human body—is particularly effective\cite{bell1994structure,vallbo1984properties}. It serves as a sensory substitution for individuals with visual, auditory, or physical disabilities\cite{kaczmarek1991electrotactile,jung2022wireless}. Also, it plays a supportive role in education, training, gaming, and social interactions\cite{choi2022transient,li2022touch}.

Tactile feedback primarily consists of electrotactile feedback \cite{vstrbac2016integrated,kaczmarek1994electrotactile} and vibrotactile feedback \cite{keef2020virtual,zhu2020development}. Electrotactile feedback offers low power consumption, minimal noise, and quick response. However, changes in impedance at the skin interface can lead to sensations such as burning or pain.\cite{jung2021skin}. In contrast, vibrotactile feedback utilizes vibrating motors that induce skin deformations and generate tactile sensations. This type of feedback is notable for its compact size, wearability, and high stability.

\begin{figure}[thbp]
	\centering
	\includegraphics[width=0.98\linewidth]{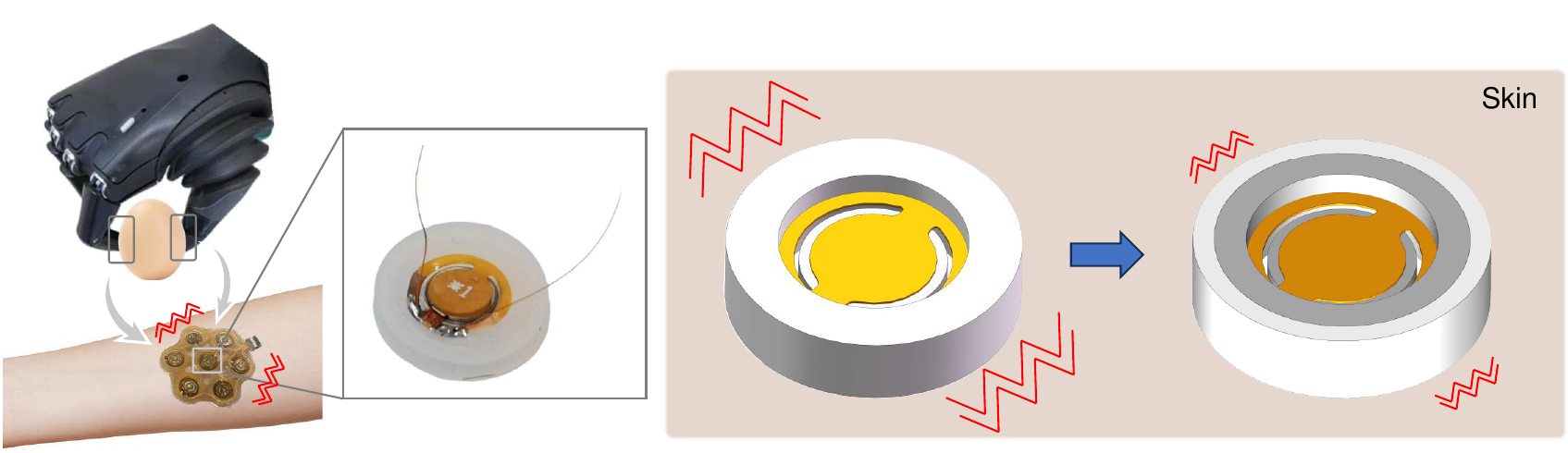}
    \captionsetup{font=footnotesize}
	\caption{The overview of the task and the proposed system. The figure illustrates how the vibrotactile interface provides feedback on the grip strength to the operator, facilitating the grasping process. The haptic units are modular components of the interface. Our aim is to put forward an optimized design of a haptic unit from the perspective of amplitude modulation.}
	\label{fig:overview}
    \vspace{-0.4cm}
\end{figure}

As shown in Figure \ref{fig:overview}, the vibrotactile interface contains several motors connected via a flexible circuit board (FPCB) and encapsulated in a silicon case. The interface has a modular structure, in which each module is referred to as a haptic unit. However, existing haptic unit designs lack amplitude modulation capabilities. Due to this limitation, the vibrotactile interface encounters the problem of mechanical crosstalk. Since both the encapsulation and the skin are elastic, the deformation caused by one unit will inevitably spread to the surrounding region, causing mechanoreceptors adjacent to other units to respond. It can interfere with the user’s ability to discern the number and location of vibrating motors, thus limiting the effective transmission of information\cite{jung2022wireless}. The problem of mechanical crosstalk necessitates the capability to rapidly attenuate the vibration amplitude at the periphery of each haptic unit.

Researchers investigated the structure of the skin and its response to vibrations at different frequencies\cite{liu2017lab,yazdi2022mechanical,moore1970survey}.
In order to simplify the analysis, the elastic model was utilized to study the propagation mechanism of vibration-induced waves in the skin\cite{kim2023mechanics,yu2019skin}. To capture the propagation and attenuation mechanisms more accurately, the viscosity property of the skin has been incorporated into the elastic model. The modified elastic model was developed, integrating an exponential term into the linear elastic framework, considering material attenuation in addition to geometric attenuation\cite{andrews2020universal}. To mitigate the mechanical crosstalk among haptic units, researchers studied the motion characteristics of common motors. The eccentric rotating mass (ERM) motor provides a larger amplitude and a smaller size compared with others\cite{kim2023mechanics}. However, arranging multiple ERM motors in a close proximity array can lead to mechanical crosstalk.

Based on the understanding of the skin and motors, researchers put forward various designs of the vibrotactile interface. The previously reported design in\cite{yu2019skin} minimized crosstalk by strategically orienting haptic units. However, this method was specific to their novel haptic units. Additionally, the design in \cite{kim2024wirelessly} incorporated a temperature module to maintain the skin at its optimal sensitivity level, thereby lowering the two-point threshold. However, this strategy led to a larger overall size of the array-based vibration interface.

\begin{figure}[!th]
    \centering
    \begin{subfigure}[a]{0.38\linewidth}
        \includegraphics[width=\linewidth]{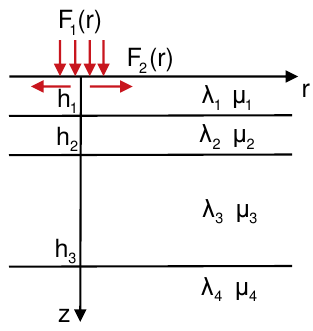}
        \captionsetup{justification=centering}
        \vspace{-0.4cm}
        \caption{}
        \label{fig:modified}
    \end{subfigure}
    \hfill
    \begin{subfigure}[a]{0.53\linewidth}
        \includegraphics[width=\linewidth]{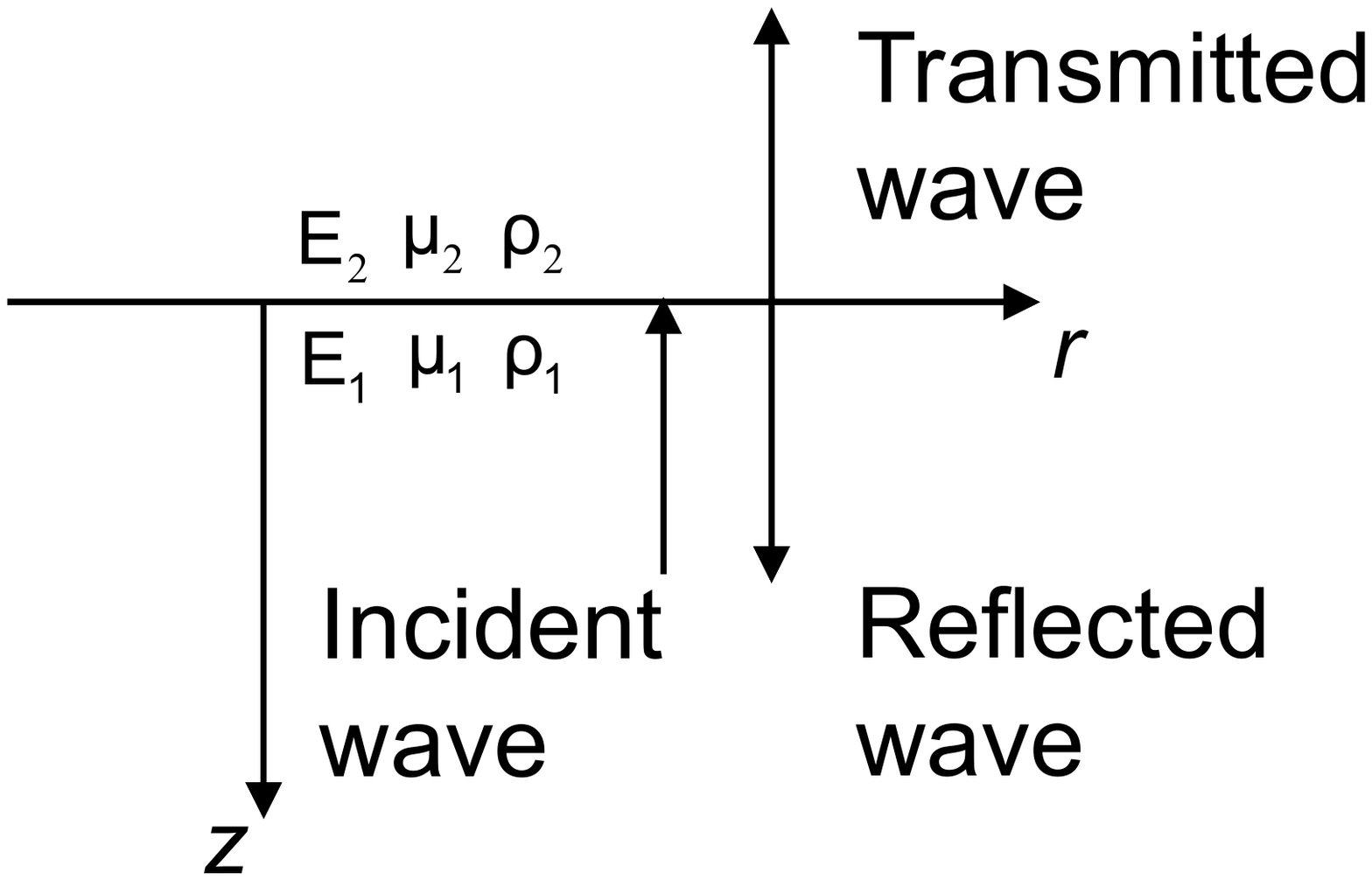}
        \captionsetup{justification=centering}
        \vspace{-0.4cm}
        \caption{}
        \label{fig:reflection}
    \end{subfigure}
    \captionsetup{font=footnotesize}
    \caption{Theoretical model. (a) The multi-layered modified elastic model in (r,z) plane. The origin is set at an arbitrary point on the contact surface between the motor and the skin. The Z-axis is oriented downward in the positive direction. (b) Illustration of the reflection and transmission of the vertically incident wave at an interface.}
    \label{fig:theory}
    \vspace{-0.4cm}
\end{figure}

This paper applied the modified elastic model to characterize the propagation and attenuation mechanisms of vibration induced by haptic units. From the perspective of amplitude modulation, optimization schemes were proposed. The performance of the optimized design was tested by a laser Doppler vibrometer (LDV), measuring the changes in amplitude compared with its previous configuration. 

Overall, our contributions are as follows:
\begin{itemize}
\item We applied the modified elastic model to simulate the multi-layer structure of the forearm. Taking viscosity into account, we theoretically analyzed the propagation and attenuation mechanisms of vibration. 
\item We changed the encapsulation of the haptic unit into a hierarchical architectural design to modulate the vibration amplitude.
\item We conducted experiments to validate the correctness of the theoretical model and the effectiveness of the proposed design.
\end{itemize}

\section{METHOD}
\label{sec:method}


\subsection{Modified Elastic Model}
\label{sec:theory}

Figure \ref{fig:modified} shows the four-layered model of the forearm, including the epidermis and dermis, hypodermis, muscle, and bone. $h_{i}$(i=1$\sim$3) represents the depth of each layer from the skin surface, while $\lambda_{i}$ (i=1$\sim$4) and $\mu_{i}$ (i=1$\sim$4) are Lame constants. The vibration propagates in the form of cylindrical waves in the epidermis and dermis in the vicinity of the motor. Therefore the analysis is conducted in a cylindrical coordinate system.  Since the distribution of force produced by the ERM motor can be approximated as uniformly distributed along the circumference\cite{moffatt2000euler}, the analysis is conducted in the $(r,z)$ plane. $\mathrm{F}_{1}$ and $\mathrm{F}_{2}$ are normal and tangential forces applied to the origin, which is an arbitrary point on the contact surface. The vibration caused by the entire motor can be regarded as the superposition of displacements generated by an infinite number of arbitrary points.

The optimized design is proposed from the perspective of changing certain parameters to achieve amplitude modulation. Therefore, the relationship between the vibration amplitude and various parameters in the model is of great importance. In order to obtain that relationship, the Navier-Stokes equations were applied for each layer of the human forearm, as shown in Equation \ref{eq:ns}. $\nabla^2$ is the Laplace operator. $\Phi$ and $\Psi$ are the scalar potential function and vector potential function of displacement. $\omega$ is the angular frequency. $v_p$ and $v_s$ are the primary wave velocity and secondary wave velocity in the medium. $\rho$ represents the density. $u_r$ and $u_z$ are displacements along the $r$ and $z$ directions. 
\begin{subequations}
\label{eq:ns}
\begin{align}
&\left(\nabla^2 + \frac{\omega^2}{v_p^2}\right) \phi = 0, \label{eq:phi} \\
&\left(\nabla^2 + \frac{\omega^2}{v_s^2}\right) \psi = 0, \label{eq:psi} \\
&v_p^2 = \frac{\lambda + 2\mu}{\rho}, \label{eq:cp} \\
&v_s^2 = \frac{\mu}{\rho}, \label{eq:cs} \\
&u_r = \frac{\partial \phi}{\partial r} + \frac{\partial^2 \psi}{\partial r \partial z}, \label{eq:ur} \\
&u_z = \frac{\partial \phi}{\partial z} - \frac{\partial^2 \psi}{\partial r^2} - \frac{1}{r} \frac{\partial \psi}{\partial r}, \label{eq:uz} \\
&\sigma_{zr} = \mu \left(\frac{\partial u_r}{\partial z} + \frac{\partial u_z}{\partial r}\right), \label{eq:sigmazr} \\
&\sigma_{zz} = \lambda\frac{1}{r} \frac{\partial}{\partial r} \left(r \frac{\partial \phi}{\partial r}\right) + \frac{\partial^2 \phi}{\partial z^2} + 2\mu \frac{\partial u_z}{\partial z}. \label{eq:sigmazz}
\end{align}
\end{subequations}

The changes of strain and stress in different layers of the forearm are continuous, resulting in the boundary conditions shown in Equation \ref{eq:boundary}, where \(k\) is wave number and \(\sigma_{zzi}\), \(\sigma_{zri}\), \(u_{ri}\), \(u_{zi}\) represent the normal stress, shear stress, displacement in the x-direction and displacement in the y-direction of layer \(i\). Substituting Equation \ref{eq:boundary} into Equation \ref{eq:ns} yields a system of equations for the vibration amplitude of the skin layers. Adding the attenuation factor $( e^{-\frac{r\eta}{E}k})$ to the linear elastic model to derive Equation \ref{eq:wave_model}.  

\begin{subequations}
\label{eq:boundary}
\begin{align}
\sigma_{zz1} \big|_{z=0} &= \mathrm{F}_1(r), \\
\sigma_{zr1} \big|_{z=0} &= \mathrm{F}_2(r), \\
\sigma_{zzi} \big|_{z=h_i} &= \sigma_{zz(i+1)} \big|_{z=h_i}, & i &= 1, 2, 3, \\
\sigma_{zri} \big|_{z=h_i} &= \sigma_{zr(i+1)} \big|_{z=h_i}, & i &= 1, 2, 3, \\
u_{ri} \big|_{z=h_i} &= u_{r(i+1)} \big|_{z=h_i}, & i &= 1, 2, 3, \\
u_{zi} \big|_{z=h_i} &= u_{z(i+1)} \big|_{z=h_i}, & i &= 1, 2, 3.
\end{align}
\end{subequations}

\begin{figure*}[th]
    \centering
    \begin{subfigure}[b]{0.15\textwidth}
        \includegraphics[width=\linewidth]{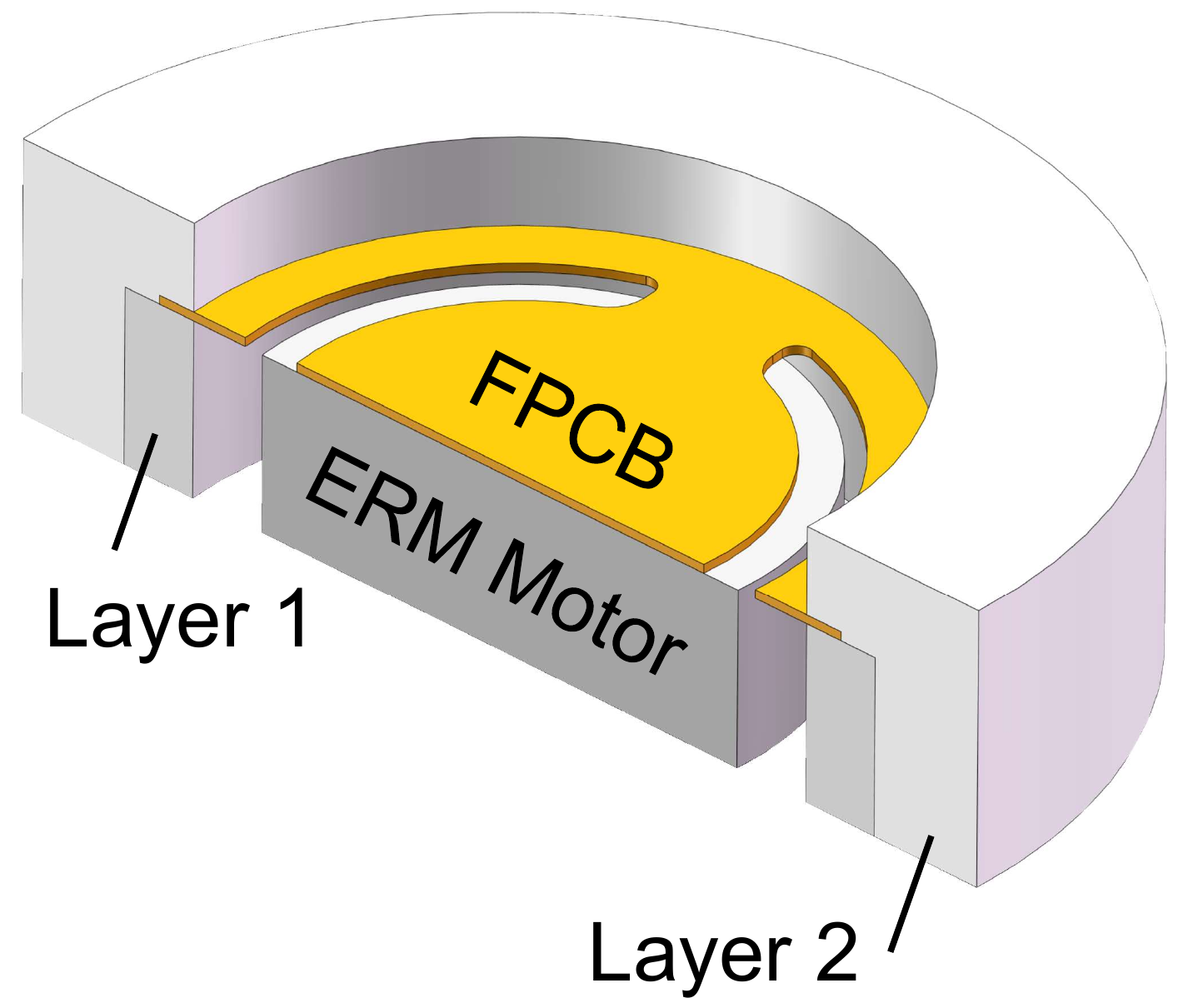}
        \captionsetup{justification=centering, font=footnotesize}
        \caption{}
        \label{fig:1-cross}
    \end{subfigure}
    \hfill
    \begin{subfigure}[b]{0.3\textwidth}
        \includegraphics[width=\linewidth]{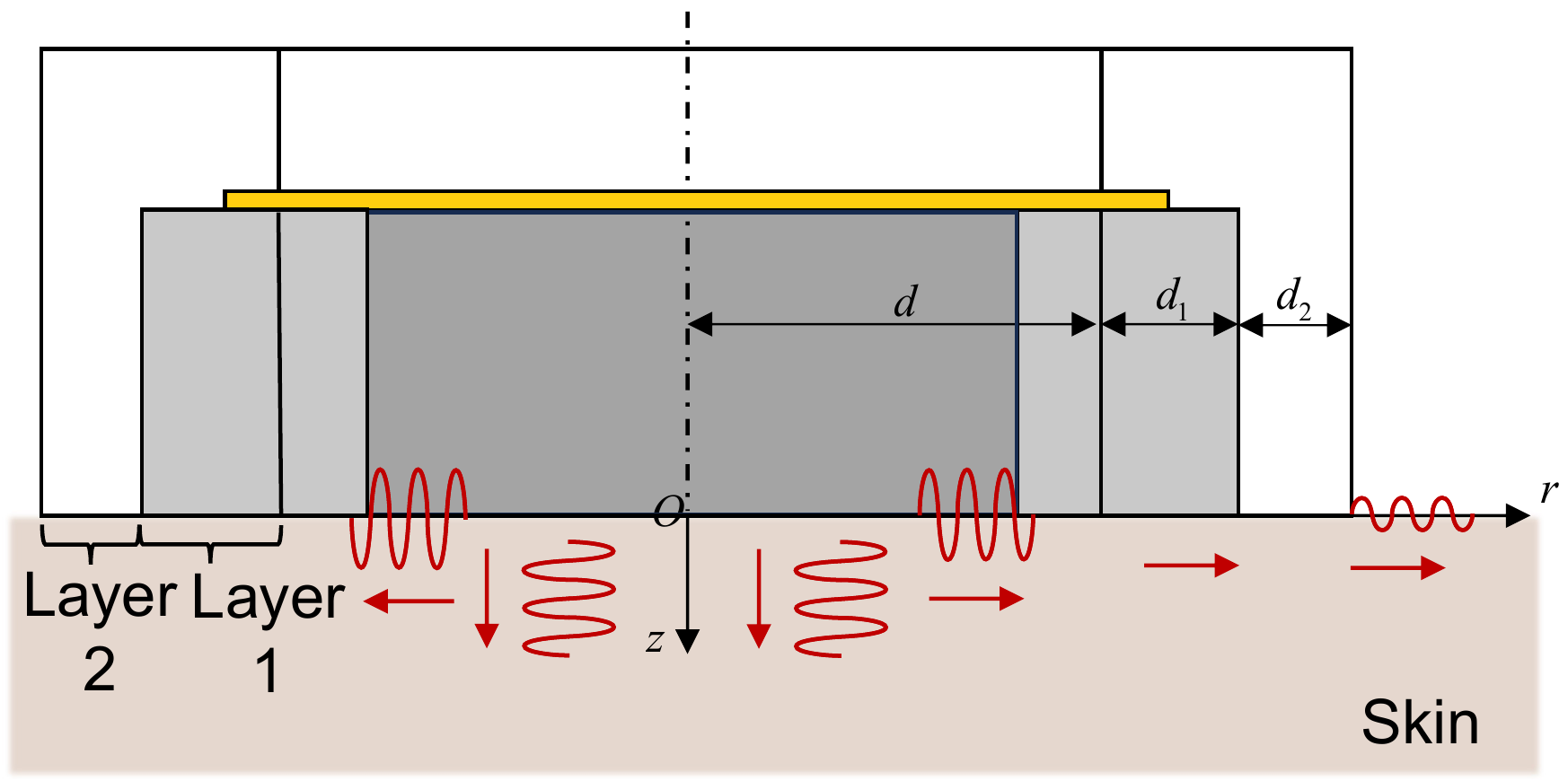}
        \captionsetup{justification=centering, font=footnotesize}
        \caption{}
        \label{fig:2-cross}
    \end{subfigure}
    \hfill
    \begin{subfigure}[b]{0.15\textwidth}
        \includegraphics[width=\linewidth]{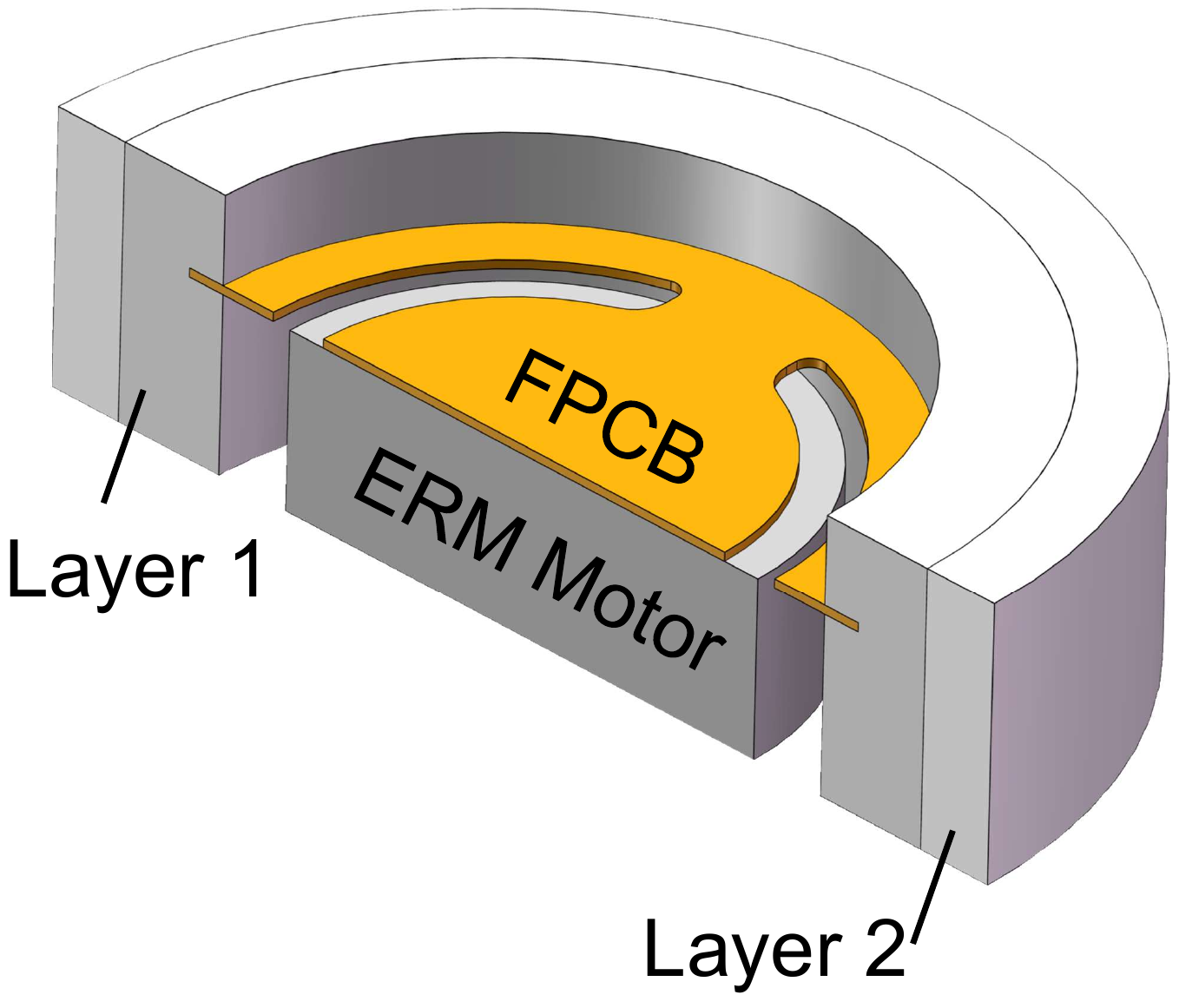}
        \captionsetup{justification=centering, font=footnotesize}
        \caption{}
        \label{fig:3-cross}
    \end{subfigure}
    \hfill
    \begin{subfigure}[b]{0.3\textwidth}
        \includegraphics[width=\linewidth]{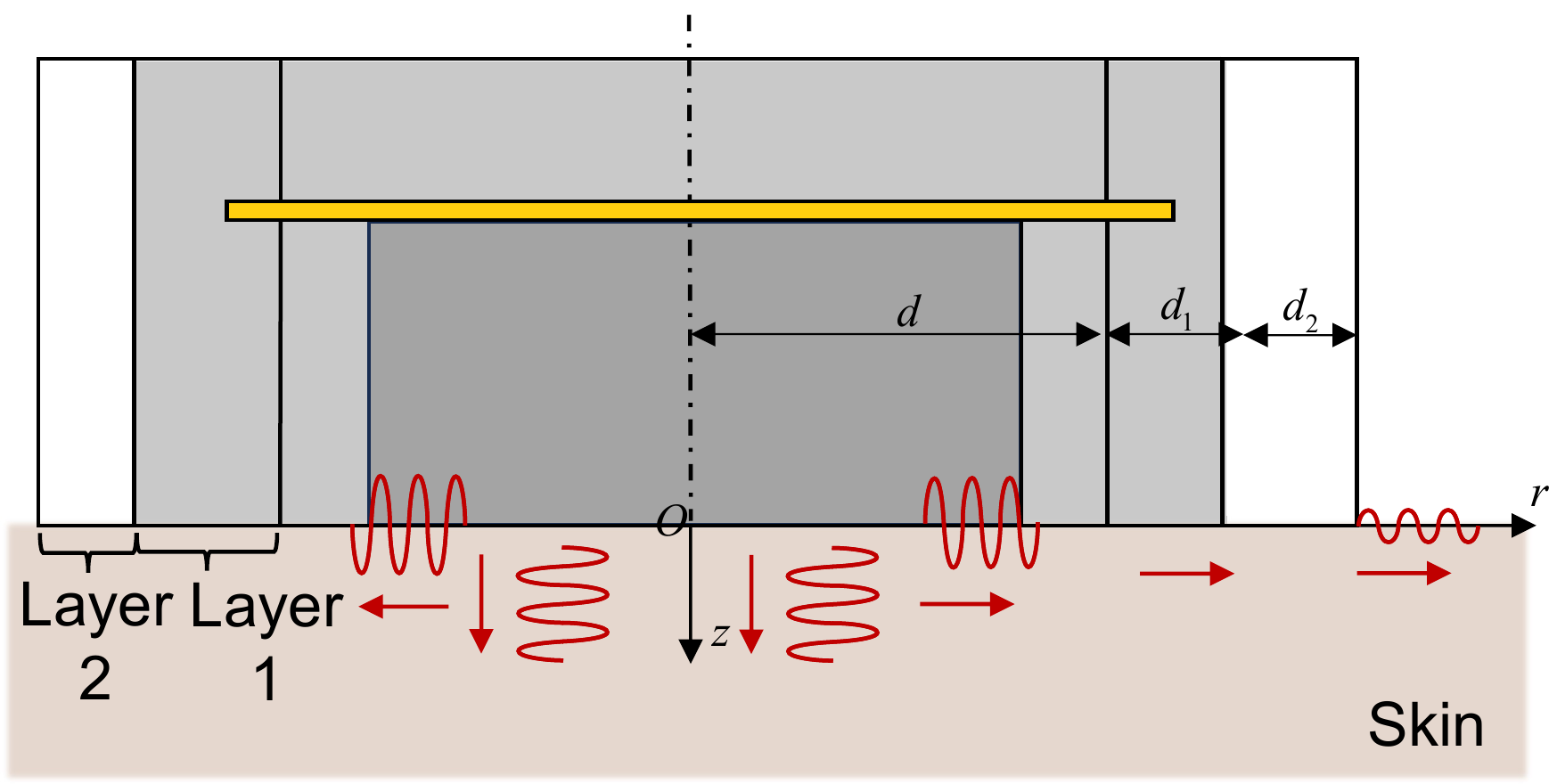}
        \captionsetup{justification=centering, font=footnotesize}
        \caption{}
        \label{fig:4-cross}
    \end{subfigure}
    \captionsetup{font=footnotesize}
    \caption{The cross-sectional view and schematic illustration depict the effect of hierarchical architectural design on amplitude attenuation for (a, b) the embedded multi-layer design, and (c, d)
the encapsulating multi-layer design. Both of them contain an ERM motor and the FPCB. The hierarchical architectural design changes the encapsulation of the haptic unit to a multi-layered structure, with Layer 1 and Layer 2 composed of different materials.}
    \vspace{-0.4cm}
    \label{fig:mechanism}
\end{figure*}

\begin{subequations}
\label{eq:wave_model}
\begin{align}
u_{ri} &= f_1(\rho_i, \mu_i, \lambda_i, z, \omega) J_1(kr) e^{-\frac{\omega \eta_i r}{E_i c_{p i}}}, \\
u_{zi} &= f_2(\rho_i, \mu_i, \lambda_i, z, \omega) J_0(kr) e^{-\frac{\omega \eta_i r}{E_i c_{s i}}}
\end{align}
\end{subequations}

In these equations, the effect of viscosity on elastic wave propagation is taken into account. \(J_0(kr)\) and \(J_1(kr)\) are solutions to the zero-order and first-order Bessel equation. They are produced when solving Equations \ref{eq:phi} and \ref{eq:psi}. $\eta_i$ and $E_i$ are the viscosity and elastic modulus of layer \(i\). $f_1$ and $f_2$ represent complex functional relationships among parameters related to the human body and the motor. 



To attenuate the vibration amplitude, the materials of the haptic unit should be viscoelastic. Therefore, Equation \ref{eq:wave_model} can also be used to analyze the propagation and attenuation mechanism of the vibration in the haptic unit, thus further guiding our design of the encapsulation for the haptic unit. 

\subsection{Hierarchical Architectural design}
\label{sec:structure}

Based on the attenuation in viscoelastic materials, a hierarchical architectural design was applied to further reduce the amplitude by reflection at the interfaces of different materials. Equation \ref{eq:reflection} is used to calculate the reflection coefficient \( k_r \) and the transmission coefficient \( k_t \) when the wave is incident perpendicularly on a surface, as shown in Figure \ref{fig:reflection}. $v_{i=\{1,2\}}$ and $\rho_{i=\{1,2\}}$ represent the wave speed and density of the incident medium and the transmitting medium.

\begin{subequations}
\label{eq:reflection}
\begin{align}
k_t &= \frac{2 \rho_1 v_1}{\rho_1 v_1 + \rho_2 v_2}, \\
k_r &= \frac{\rho_1 v_1 - \rho_2 v_2}{\rho_1 v_1 + \rho_2 v_2}.
\end{align}
\end{subequations}

Figure \ref{fig:1-cross} and \ref{fig:3-cross} display the embedded multi-layer design and encapsulating multi-layer design. Both of them are two-layered structures, and the thickness of each layer is 1.25 mm. The overall diameter of the haptic unit is 14mm for both designs. In the embedded multi-layer design, the flexible circuit board is in contact with both layers, while in the encapsulating multi-layer design, it only contacts Layer 1. The schematic diagrams illustrating the mechanism of hierarchical structure and its role in causing amplitude attenuation are shown in Figure \ref{fig:2-cross} and \ref{fig:4-cross}.

Compared with a single-layer design of the same size, the vibration amplitude at the same location on the skin surface is smaller in the hierarchical architectural design, as shown in Constraint \ref{eq:constraint_ini}. 

\begin{subequations}
    \begin{align}
        &  f_1 J_1(kr) k_{t_1} k_{t_2} e^{\frac{-\omega \eta _1 d}{E_1 v_{p_1}}+\frac{-\omega \eta _{l_1}d_1}{E_{l_1}v_{p_{l_1}}}+\frac{-\omega \eta _{l_2} d_2}{E_{l_2}v_{p_{l_2}}}+\frac{-\omega \eta _1 (r-d-d_1-d_2)}{E_1{v_{p_1}}}} < \notag \\
        & f_1 J_1(kr)k_t e^{\frac{-\omega \eta _1 d}{E_1v_{p_1}}+\frac{-\omega \eta _1 (r-d-d_1-d_2)}{E_1v_{p_1}}+\frac{-\omega \eta _c (d_1+d_2)}{E_cv_{p_c}}} \\
        &  f_2 J_o(kr) k_{t_1} k_{t_2} e^{\frac{-\omega \eta _1 d}{E_1 v_{s_1}}+\frac{-\omega \eta _{l_1}d_1}{E_{l_1}v_{s_{l_1}}}+\frac{-\omega \eta _{l_2} d_2}{E_{l_2}v_{s_{l_2}}}+\frac{-\omega \eta _1 (r-d-d_1-d_2)}{E_1{v_{s_1}}}} < \notag \\
        & f_2 J_o(kr)k_t e^{\frac{-\omega \eta _1 d}{E_1v_{s_1}}+\frac{-\omega \eta _1 (r-d-d_1-d_2)}{E_1v_{s_1}}+\frac{-\omega \eta _c (d_1+d_2)}{E_cv_{s_c}}}
    \end{align}
    \label{eq:constraint_ini}
\end{subequations}

As shown in Figure \ref{fig:mechanism}, $d$, $d_1$, and $d_2$ are the inner radius of Layer 1, the thickness of Layer 1, and the thickness of Layer 2. $k_{t_1}$ and $k_{t_2}$ are parameters in the hierarchical architectural design, which represent the transmission coefficients at the interfaces between Layer 1 and Layer 2, and between Layer 2 and the air, respectively. $k_{t_c}$ is the transmission coefficients at the interfaces between the one-layer structure in the one-layer design and the air. $E$, $\eta$, $v_p$, and $v_s$ represent the elastic modulus, viscosity, the wave speed of the primary wave, and the wave speed of the secondary wave, respectively. Subscripts 1, $l_1$, $l_2$, and $c$ represent the epidermis and dermis, Layer 1, Layer 2, and the material of the one-layer design.
 

By eliminating the common terms on both sides of Constrain \ref{eq:constraint_ini} and substituting Equation \ref{eq:cp} and Equation \ref{eq:cs} into Constraint \ref{eq:constraint_ini}, Constraint \ref{eq:constraints} can be derived as follows. $\nu_{l_1}$, $\nu_{l_2}$, and $\nu_c$ are the Poisson's ratio of Layer 1, Layer 2, and the material of one-layer design. $\rho_{l_1}$, $\rho_{l_2}$ and $\rho_c$ are the density of Layer 1, Layer 2, and the material of one-layer design.

{\small
\begin{subequations}
\label{eq:constraints}
\begin{align}
&\frac{\eta_{l_1} d_1 \sqrt{(1+\nu_{l_1})(1-2 \nu_{l_1}) \rho_{l_1}}}{E_{l_1} \sqrt{E_{l_1} (1-\nu_{l_1})}} + \frac{\eta_{l_2} d_2 \sqrt{(1+\nu_{l_2})(1-2 \nu_{l_2}) \rho_{l_2}}}{E_{l_2} \sqrt{E_{l_2} (1-\nu_{l_2})}} \notag \\
&> \frac{\eta_c (d_1 + d_2) \sqrt{(1+\nu_c)(1-2 \nu_c) \rho_c}}{E_c \sqrt{E_c (1-\nu_c)}}, \\
&\frac{\eta_{l_1} d_1 \sqrt{2(1+\nu_{l_1}) \rho_{l_1}}}{E_{l_1} \sqrt{E_{l_1}}} + \frac{\eta_{l_2} d_2 \sqrt{2(1+\nu_{l_2}) \rho_{l_2}}}{E_{l_2} \sqrt{E_{l_2}}}  \notag \\
&> \frac{\eta_c (d_1 + d_2) \sqrt{2(1+\nu_c) \rho_c}}{E_c \sqrt{E_c}}, \\
&\frac{E_{l_1} \rho_{l_1} (1-\nu_{l_1})}{(1+\nu_{l_1}) (1-2 \nu_{l_1})} < \frac{E_{l_2} \rho_{l_2} (1-\nu_{l_2})}{(1+\nu_{l_2}) (1-2 \nu_{l_2})}, \\
&\frac{E_{l_1} \rho_{l_1}}{1+\nu_{l_1}} < \frac{E_{l_2} \rho_{l_2}}{1+\nu_{l_2}}.
\end{align}
\end{subequations}
}

To achieve quicker attenuation of vibration amplitude under identical vibration excitation conditions, the amplitude around the optimized design should be smaller compared to the original, indicating a larger gap on both sides of the inequality sign. Therefore, it is necessary to reduce the elastic modulus of Layer 1 and Layer 2 while increasing their density, viscosity, and Poisson's ratio. Additionally, the elastic modulus and density of Layer 1 should be lower than those of Layer 2.

\section{EXPERIMENTS}
\label{sec:experiment}
\subsection{Experimental Setup}

\subsubsection{Experiment Platform}

\begin{figure}[thbp]
	\centering
	\includegraphics[width=0.75\linewidth]{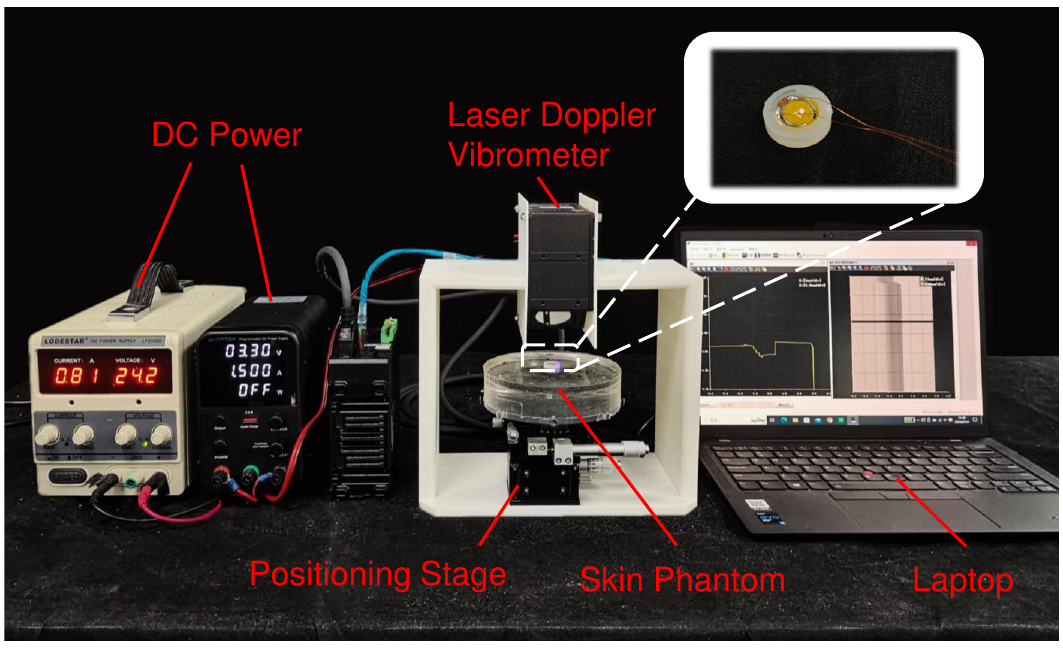}
    \captionsetup{font=footnotesize}
	\caption{Photo of the experimental setup for measuring the amplitude of vibration propagating through the skin phantom.}
	\label{fig:device}
    \vspace{-0.4cm}
\end{figure}

Figure \ref{fig:device} illustrates the experimental platform used for measuring vibration amplitude. The haptic unit was encapsulated and attached to a multi-layer skin phantom. To reduce any interference with the motion of the haptic module, 0.1 mm enameled wires were used. To compare the modulation effects of these two designs under the same excitation, both schemes utilized an LCM0720A3463F ERM motor (Leader, Inc.) with a diameter of 7mm.

The ERM motor within the haptic unit vibrated at a frequency of 100-150 Hz, inducing deformation in the skin phantom. The vibration traveled through the phantom and the layered structure within the haptic unit. The LDV (Keyence, Inc.) was used to measure the vibration amplitude at various points. For an accurate description of the displacement-time curve, a sampling rate of 2 kHz was employed. To enable precise adjustments in the X, Y, and R directions, the phantom was mounted on a micrometer positioning stage.

To eliminate interference from low-frequency noise and power-line frequency noise, the collected data were filtered using a Butterworth filter with a frequency range of 40 to 200 Hz, together with a 50 Hz comb filter. To demonstrate the performance of the hierarchical architectural design, the ERM motor's amplitude was used as a reference. The normalized amplitude was calculated for units with different designs and materials to evaluate the effectiveness of the optimized design. To improve the reliability and accuracy of data, each experiment was repeated three times.


\subsubsection{Optimized Designs and Evaluation}
Considering the wearability of haptic units, silicone materials such as Ecoflex and Dragon Skin (Smooth-On, Inc.) were commonly used for encapsulation. This paper selected materials from common silicone types to construct single-layer and hierarchical structure haptic units of the same size. The single-layer unit served as the control group, while the hierarchically designed unit, using the same material as the control group, formed the experimental group.

Given that the density, Poisson's ratio, and viscosity of different silicone materials do not differ significantly, the elastic modulus of the layers is the primary factor determining the effectiveness of amplitude modulation, as indicated by Equation \ref{eq:constraints}. To quantify this, we used a tensile and compression testing platform (MARK-10, Inc.) to determine the stress-strain curves of various silicone types at the operating frequency of the ERM motor.


\subsection{Elastic Modulus of Different Materials}
\label{exp:elas}
Figure \ref{fig:E} illustrates the stress-strain curve of different types of silicone. To highlight the advantages of the hierarchical structure over the control group, we selected three materials with significantly different elastic modulus. By applying these three silicones at different layers in the hierarchical structure, the effects of different materials on the amplitude modulation of the haptic unit can be studied.

Within the strain range induced by the ERM motor, Ecoflex 00-10 exhibits the lowest elastic moduli of 15 kPa, while Dragon Skin 30 reaches the highest of 355 kPa. Dragon Skin 10 Medium falls between these extremes with a modulus of 250 kPa. The haptic units using only Ecoflex 00-10, Dragon Skin 10 Medium, and Dragon Skin 30 serve as the control groups.

\subsection{Performance of Optimized Design}
\label{exp:comp}
\begin{figure*}[t]
    \centering
    \begin{subfigure}[b]{0.3\linewidth}
        \includegraphics[width=\linewidth]{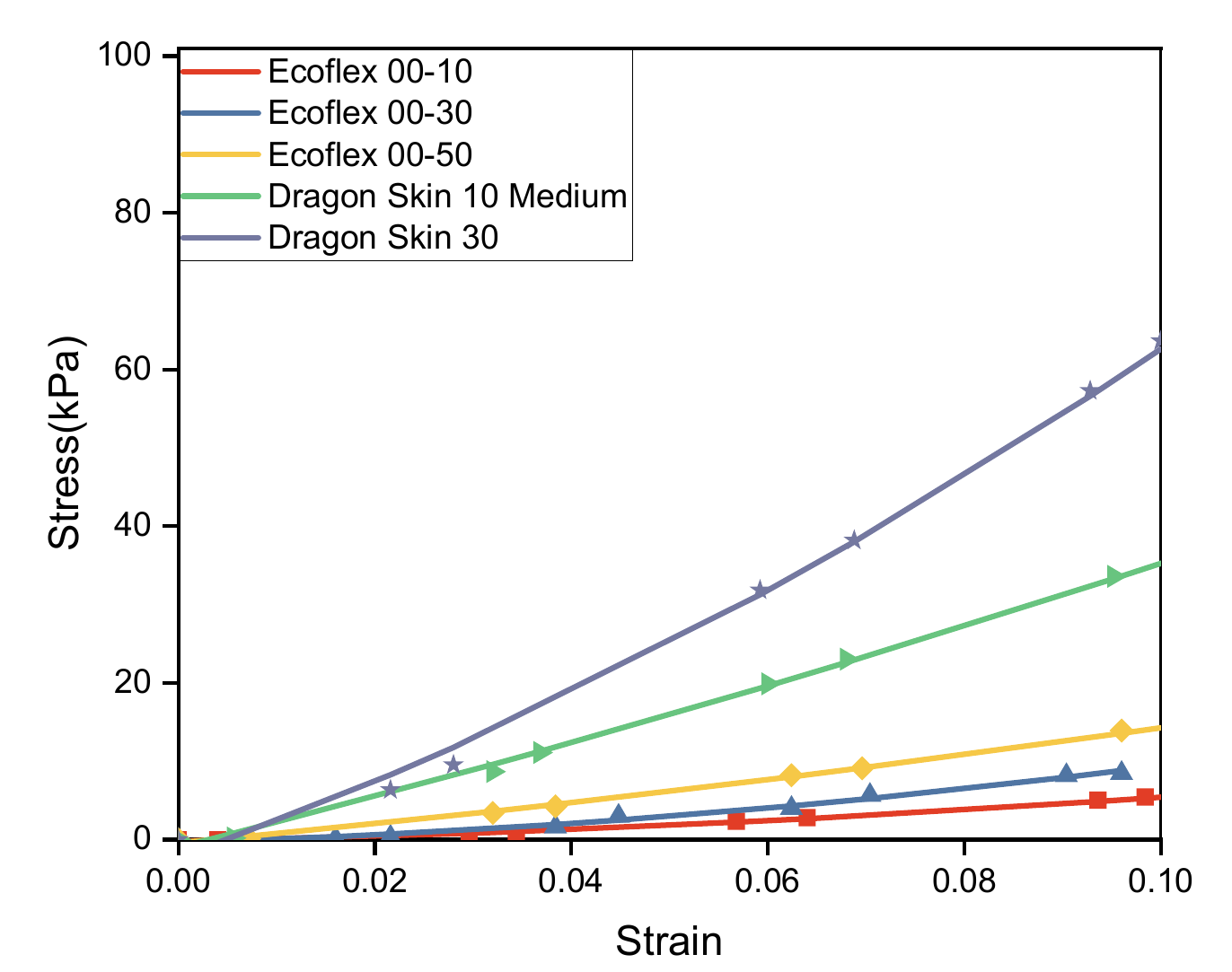}
        \captionsetup{justification=centering, font=footnotesize}
        \caption{}
        \label{fig:E}
    \end{subfigure}
    \hfill
    \begin{subfigure}[b]{0.3\linewidth}
        \includegraphics[width=\linewidth]{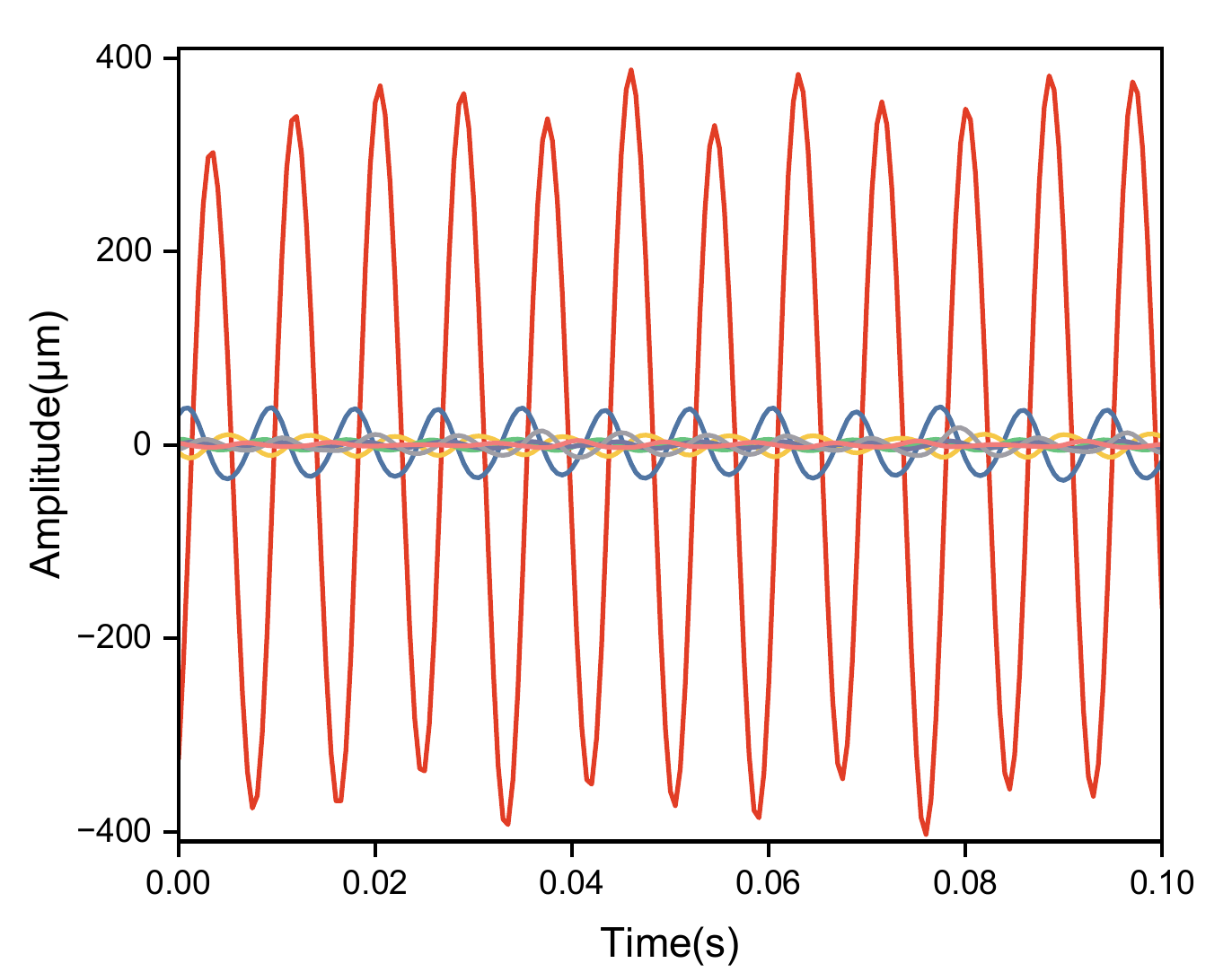}
        \captionsetup{justification=centering, font=footnotesize}
        \caption{}
        \label{fig:amplitude-a}
    \end{subfigure}
    \hfill
    \begin{subfigure}[b]{0.3\linewidth}
        \includegraphics[width=\linewidth]{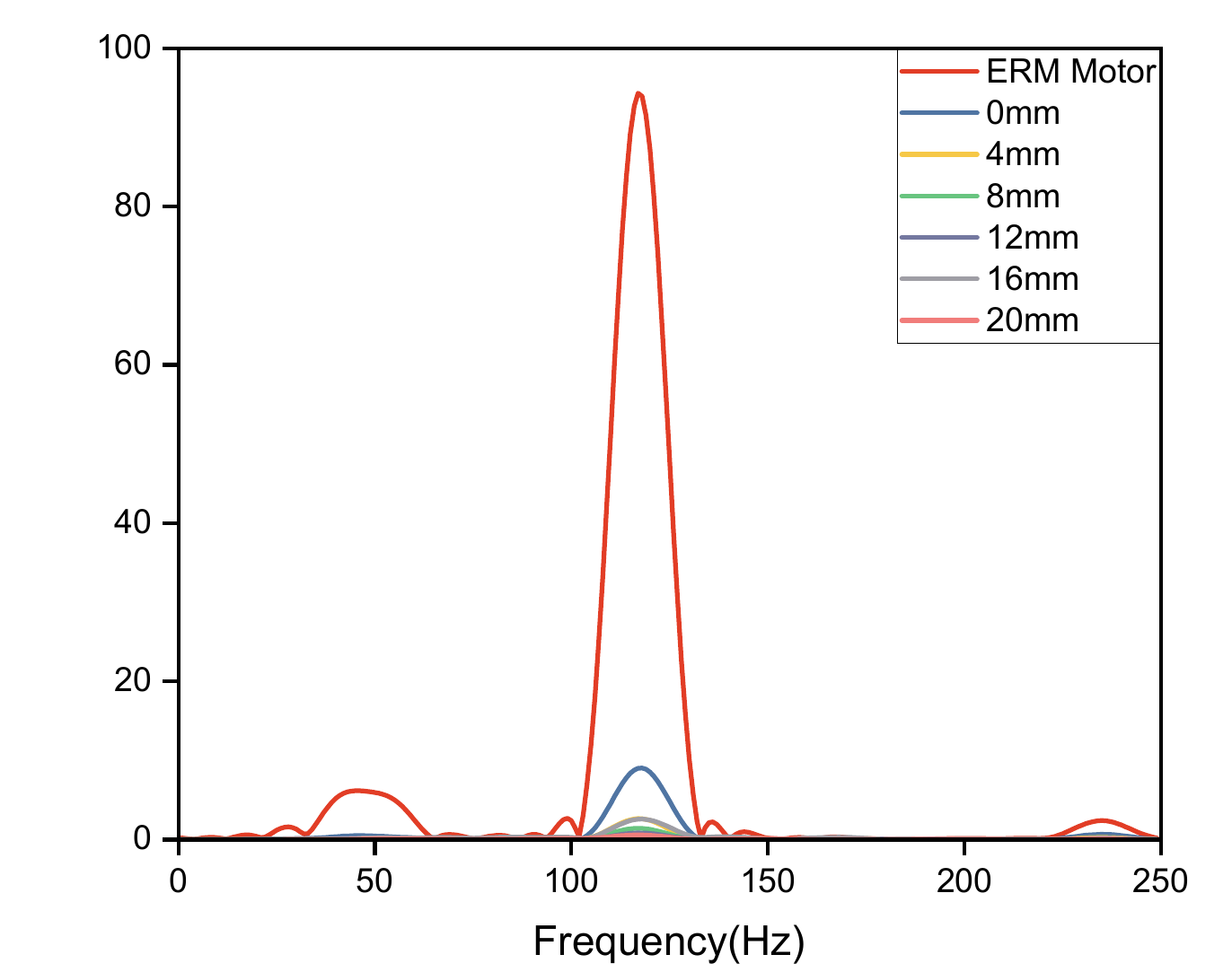}
        \captionsetup{justification=centering, font=footnotesize}
        \caption{}
        \label{fig:amplitude-b}
    \end{subfigure}
    \captionsetup{font=footnotesize}
    \caption{The sub-figures (from left to right) illustrate: (a) the stress-strain curve of Ecoflex 00-10, Ecoflex 00-30, Ecoflex 00-50, Dragon Skin 10 Medium, and Dragon Skin 30. (b) the vibration amplitude-time curve and (c) the frequency of a hierarchical architectural design at various radial distances on the skin phantom. The materials for Layer 1 and Layer 2 are Ecoflex 00-10 and Dragon Skin 30. The boundary of the haptic unit is set at 0 mm.}
    \label{fig:stress-strain-and-amplitude}
\end{figure*}

Figure \ref{fig:amplitude-a} illustrates the displacements induced by a haptic unit with a hierarchical architectural design at different locations of the skin phantom. The amplitude of the ERM motor is approximately 300 $\upmu$m. The spectrum is presented in \ref{fig:amplitude-b}. The waveform of vibration induced by other haptic units on the skin phantom is similar to those shown in Figure \ref{fig:amplitude-a}, and the maximum value of the spectrum occurs at the same position as Figure \ref{fig:amplitude-b}. The main difference is the vibration amplitude at the edge of the haptic units. The normalized amplitude, measured when the distance is zero, is applied to indicate the attenuation of vibration along the skin. 

\begin{figure*}[t]
    \centering
    \begin{subfigure}[b]{0.245\linewidth}
        \includegraphics[width=\linewidth]{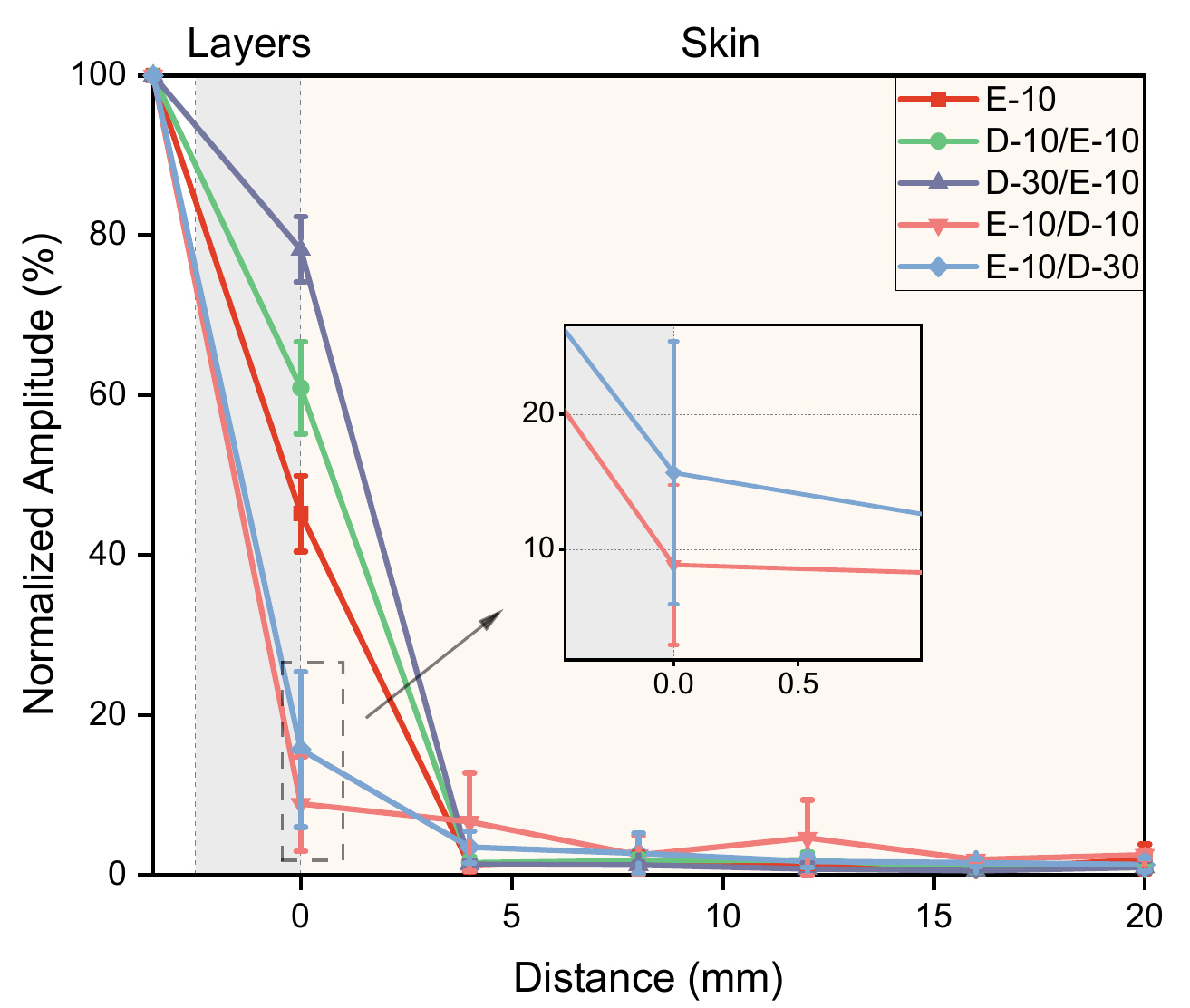}
        \captionsetup{justification=centering, font=footnotesize}
        \caption{}
        \label{fig:amplitude-c}
    \end{subfigure}
    \hfill
    \begin{subfigure}[b]{0.245\linewidth}
        \includegraphics[width=\linewidth]{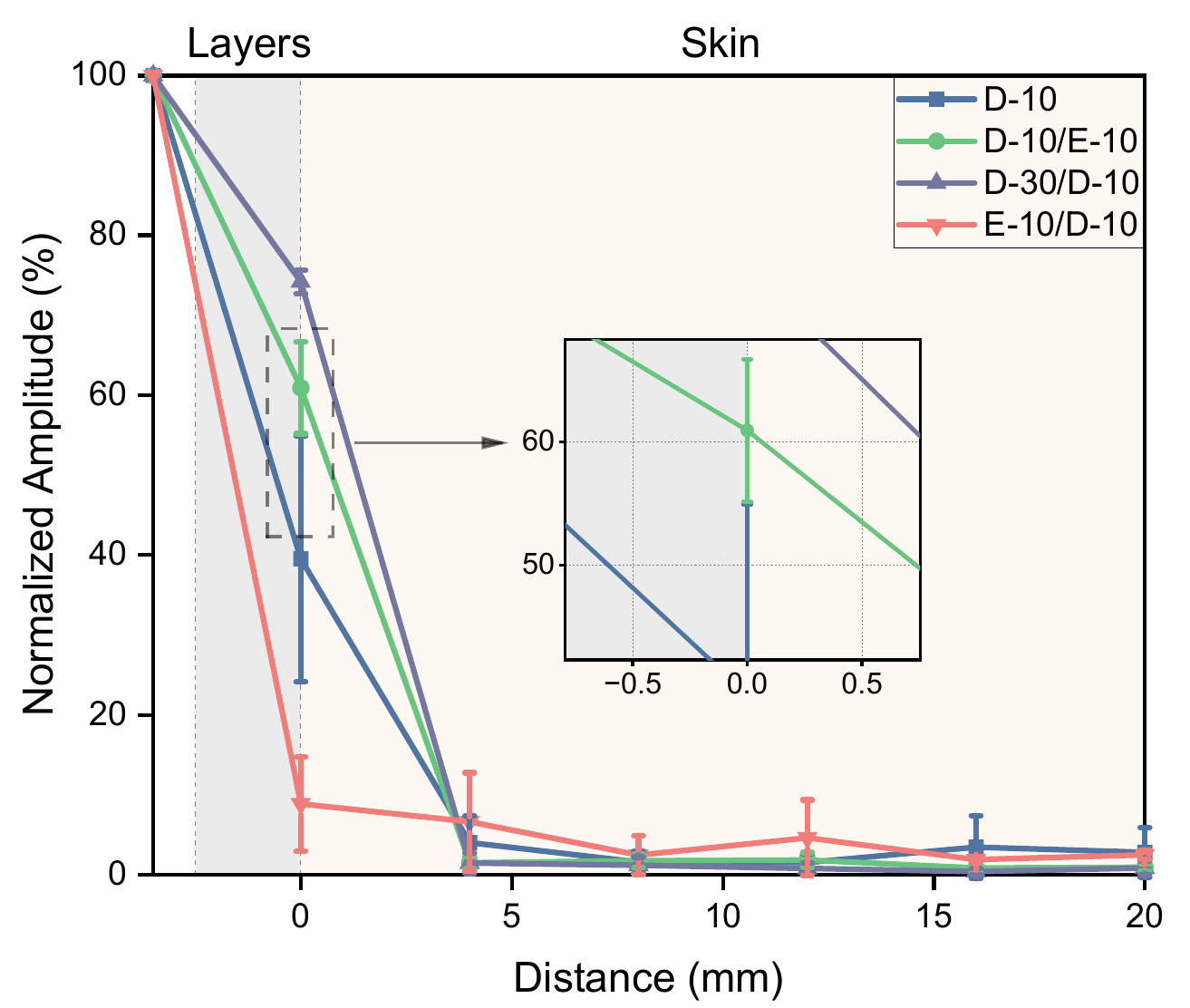}
        \captionsetup{justification=centering, font=footnotesize}
        \caption{}
        \label{fig:amplitude-d}
    \end{subfigure}
    \hfill
    \begin{subfigure}[b]{0.245\linewidth}
        \includegraphics[width=\linewidth]{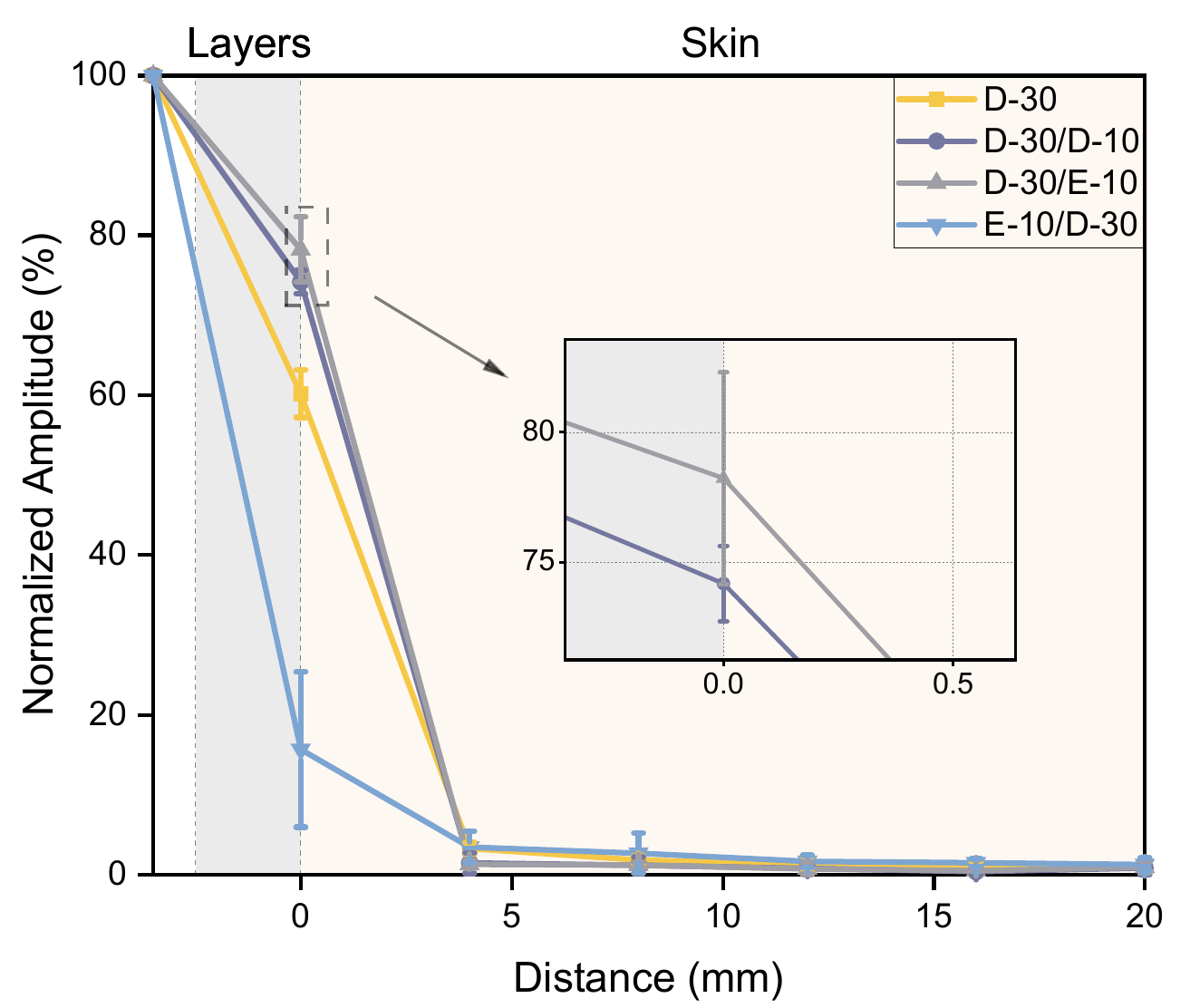}
        \captionsetup{justification=centering, font=footnotesize}
        \caption{}
        \label{fig:amplitude-e}
    \end{subfigure}
    \hfill
    \begin{subfigure}[b]{0.245\linewidth}
        \includegraphics[width=\linewidth]{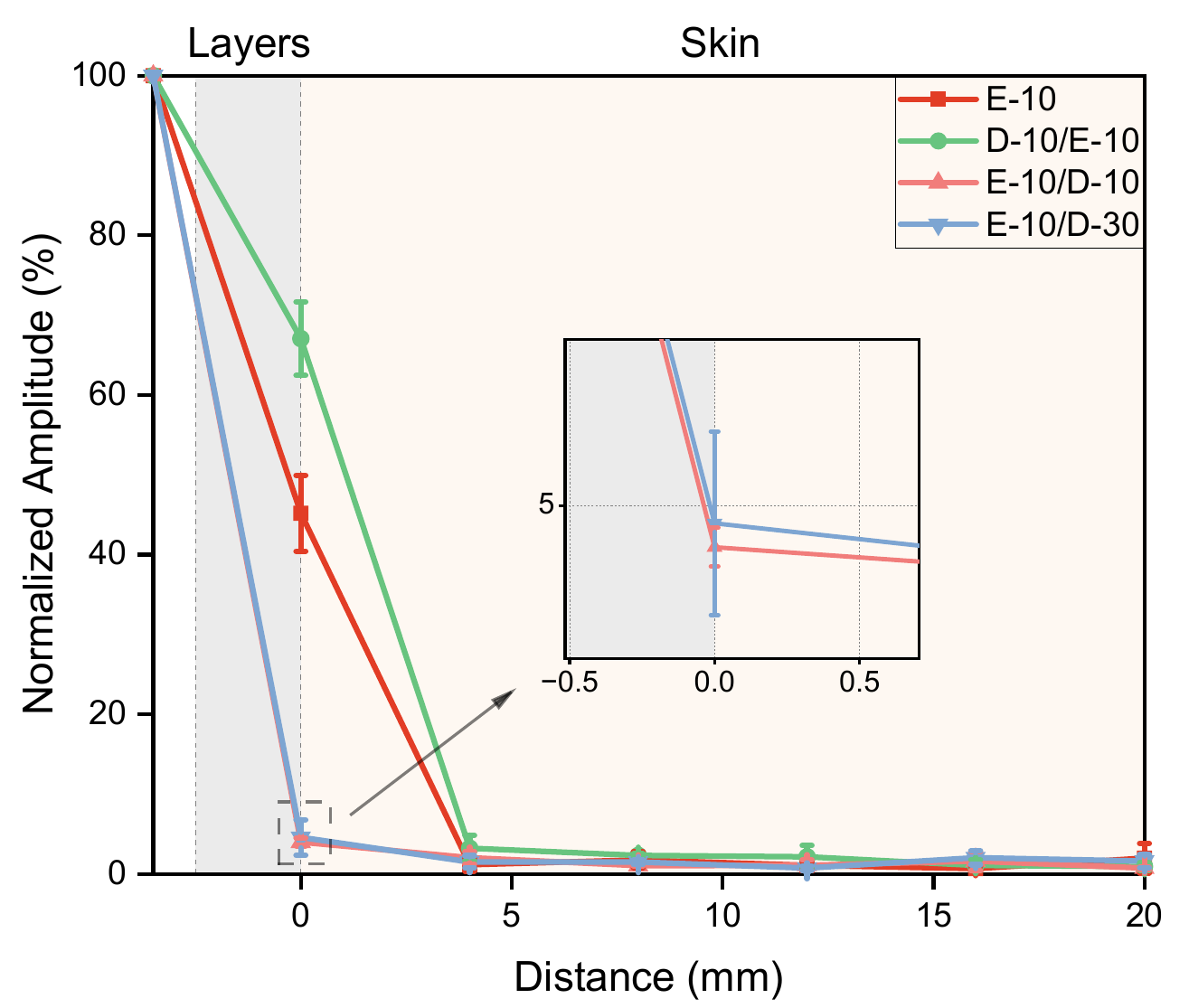}
        \captionsetup{justification=centering, font=footnotesize}
        \caption{}
        \label{fig:amplitude-f}
    \end{subfigure}
    \hfill
    \begin{subfigure}[b]{0.245\linewidth}
        \includegraphics[width=\linewidth]{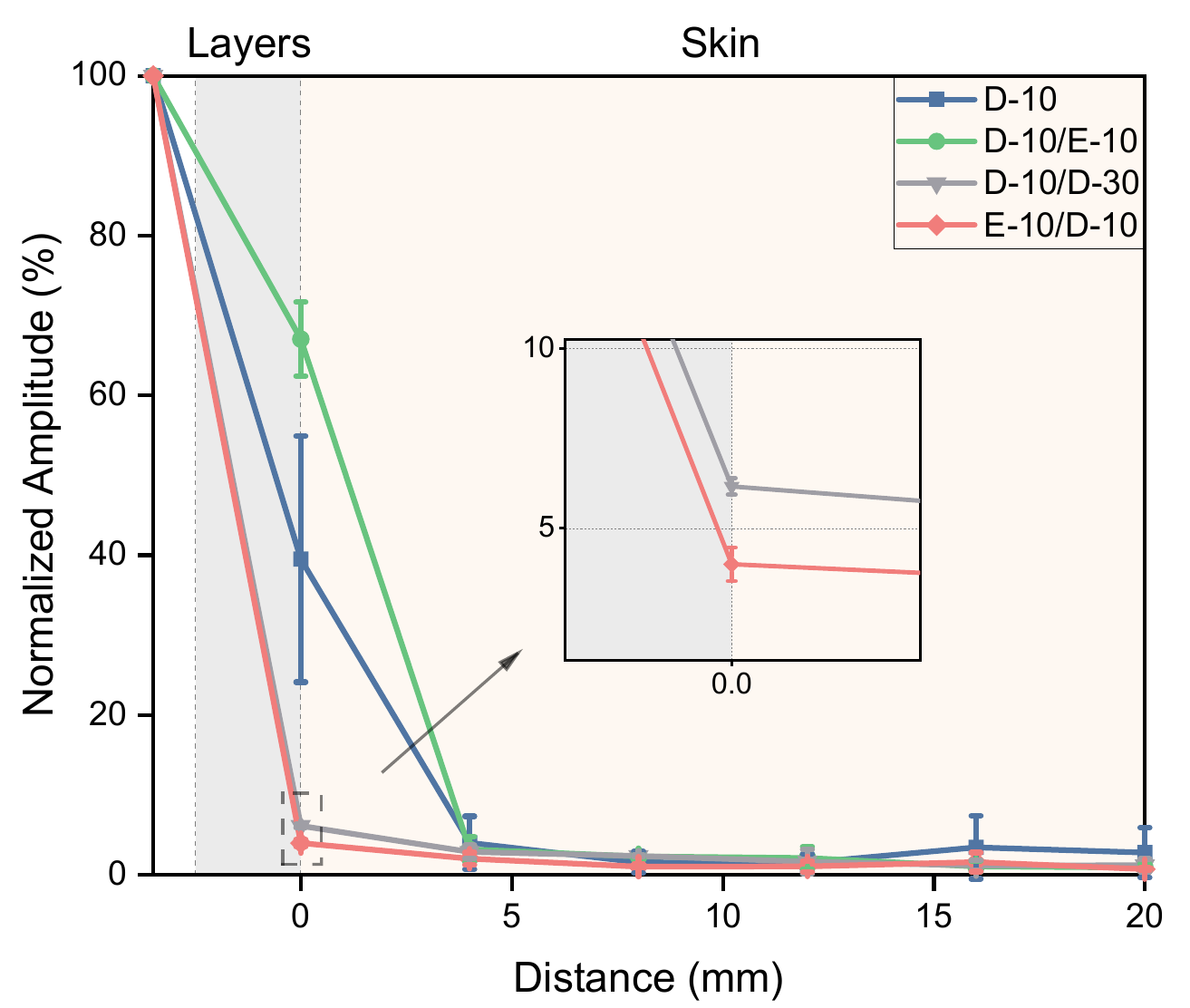}
        \captionsetup{justification=centering, font=footnotesize}
        \caption{}
        \label{fig:amplitude-g}
    \end{subfigure}
    \hfill
    \begin{subfigure}[b]{0.245\linewidth}
        \includegraphics[width=\linewidth]{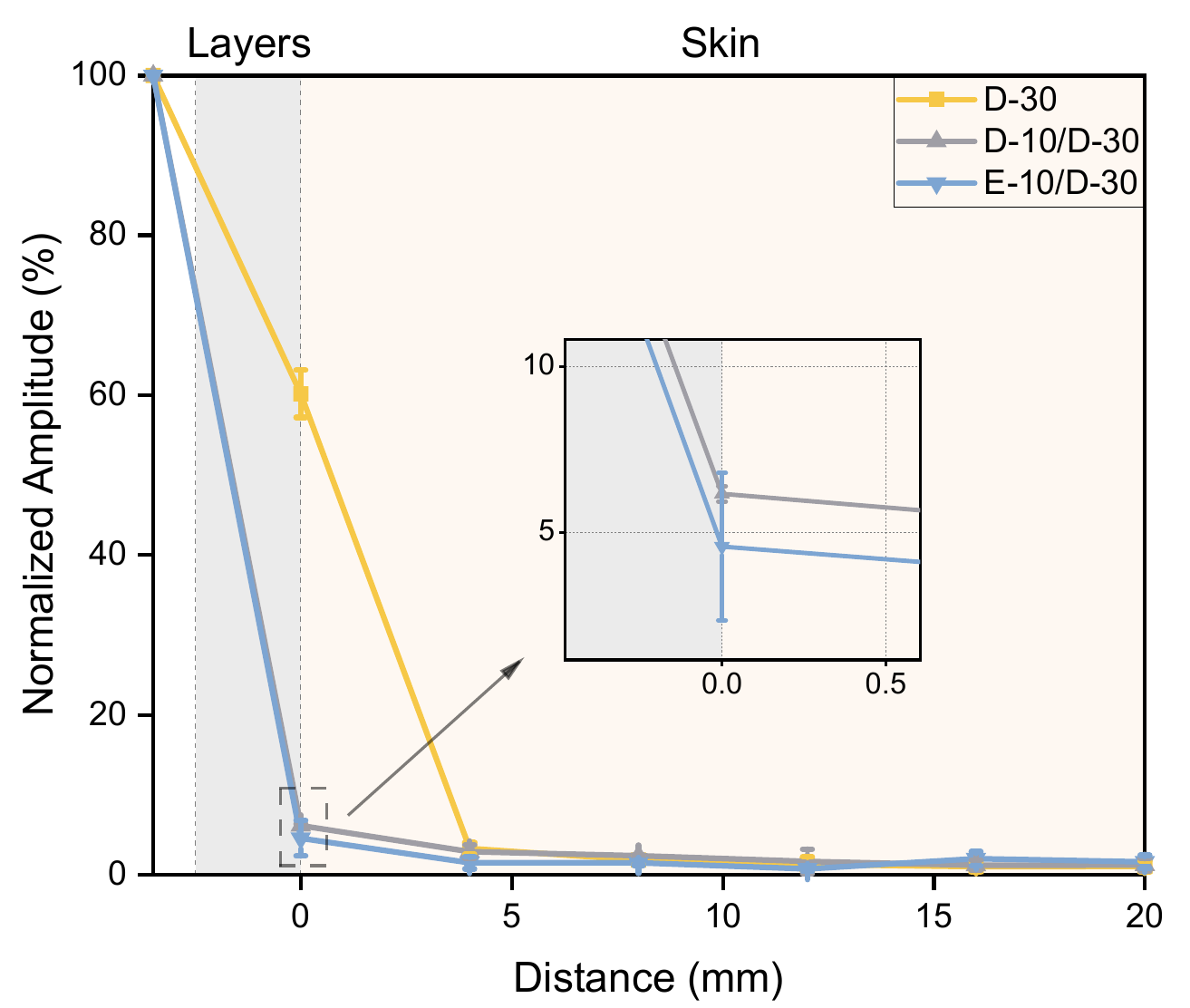}
        \captionsetup{justification=centering, font=footnotesize}
        \caption{}
        \label{fig:amplitude-h}
    \end{subfigure}
        \begin{subfigure}[b]{0.245\linewidth}
        \includegraphics[width=\linewidth]{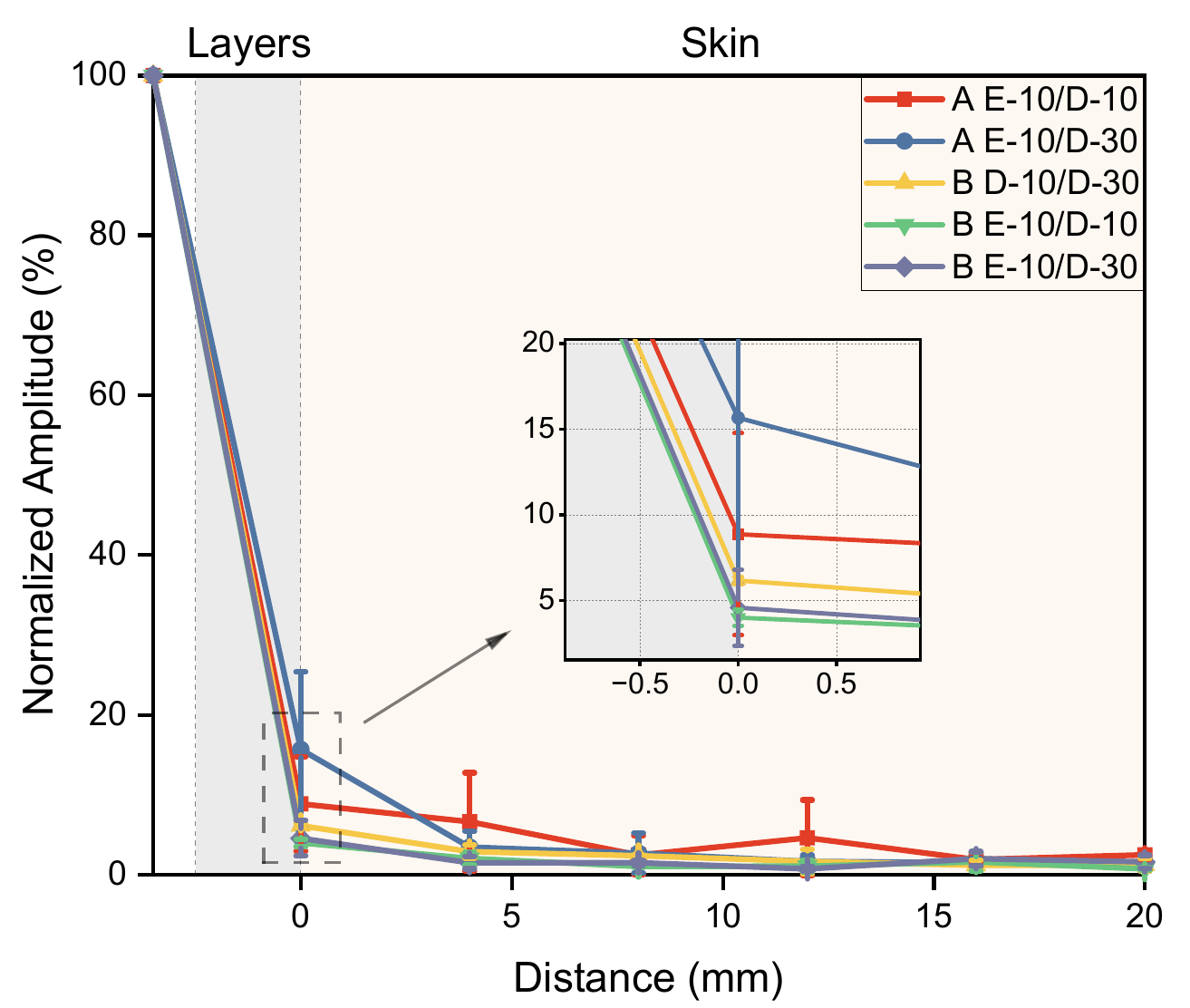}
        \captionsetup{justification=centering, font=footnotesize}
        \caption{}
        \label{fig:amplitude-i}
    \end{subfigure}
    \hfill
    \begin{subfigure}[b]{0.245\linewidth}
        \includegraphics[width=\linewidth]{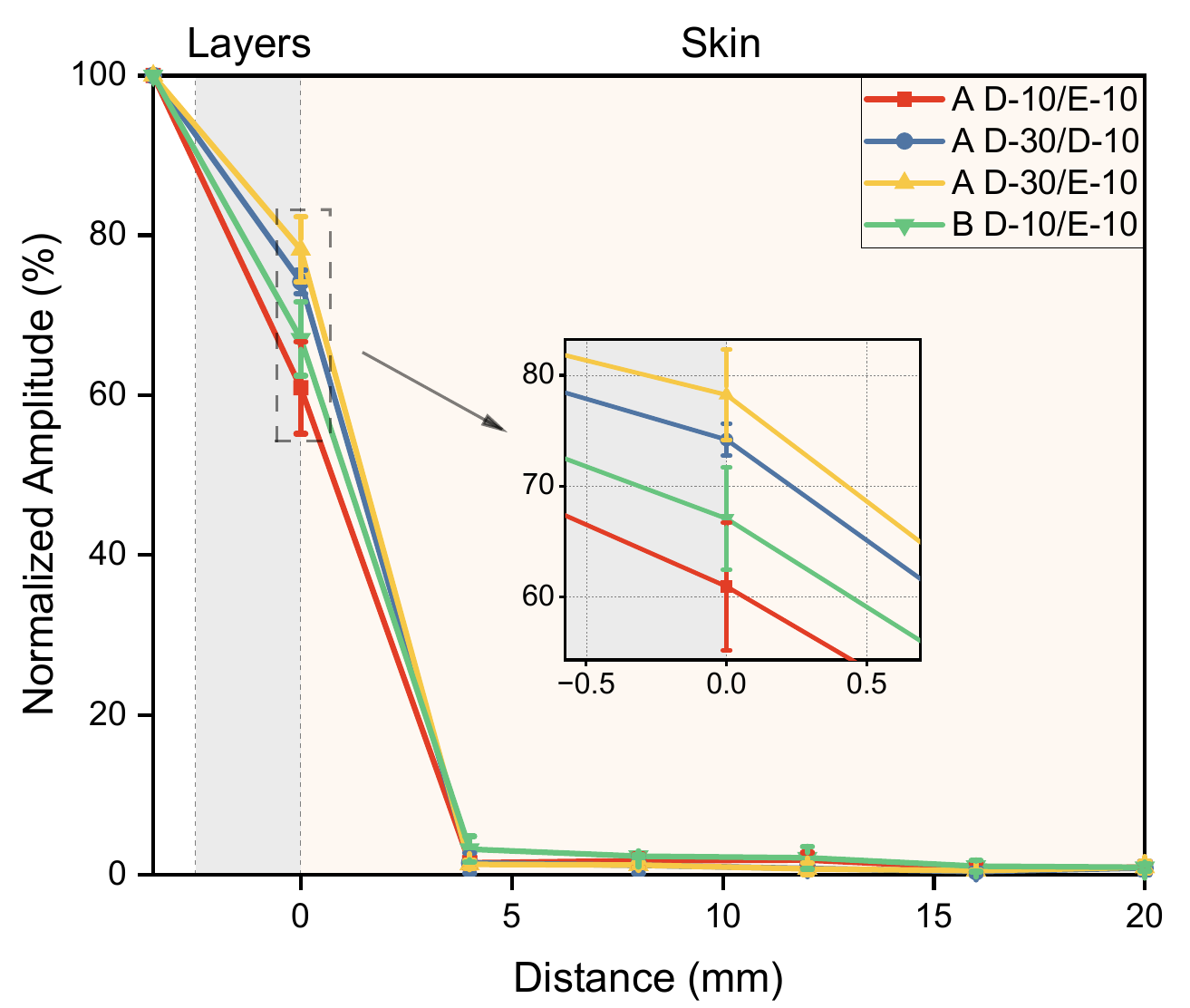}
        \captionsetup{justification=centering, font=footnotesize}
        \caption{}
        \label{fig:amplitude-j}
    \end{subfigure}
    \captionsetup{font=footnotesize}
    \caption{The diagrams illustrate the normalized amplitude at different radial distances on the skin phantom measured by the LDV. (a), (b), and (c) are results for the embedded multi-layer design, while (d), (e), and (f) are for the encapsulating multi-layer design. The materials adopted for each unit are presented in the form of Layer 1/Layer 2. The haptic units using only Ecoflex 00-10 (E-10), Dragon Skin 10 Medium (D-10), and Dragon Skin 30 (D-30) serve as the control groups. The control group and the experimental group containing the same materials are compared in the same figure. The layers and skin regions of the haptic units are shaded. The edge of the haptic unit is set at 0mm. The error bars represent the standard deviation of the normalized amplitude. (f) The normalized amplitude of the embedded multi-layer design and the encapsulating multi-layer design with different materials for amplitude attenuation. (g) The normalized amplitude of the embedded multi-layer design and encapsulating multi-layer design with different materials for amplitude attenuation enhancement.}
    \label{fig:amplitude}
    \vspace{-0.4cm}
\end{figure*}

Figure \ref{fig:amplitude} illustrates the amplitude of the control groups and the experimental group with a hierarchical architectural structure. Experimental groups containing the same materials as the control groups are presented in the same figure to demonstrate the effectiveness of amplitude modulation. When Ecoflex 00-10, Dragon Skin 10 Medium, and Dragon Skin 30 are used for the control group, the normalized amplitude at the periphery of the haptic unit is 45.18\%, 39.5\%, and 60.18\%, respectively. The amplitude of vibrations from both designs exhibits a quasi-exponential decay along the radial direction of the skin, consistent with the derivations from the theoretical model. Figure \ref{fig:amplitude-c}, \ref{fig:amplitude-d}, and \ref{fig:amplitude-e} illustrate the results for the embedded multi-layer design. Figure \ref{fig:amplitude-f}, \ref{fig:amplitude-g}, and \ref{fig:amplitude-h} are the results for encapsulating multi-layer design. 

Haptic units with hierarchical architectural design show greater amplitude attenuation capability. When using Ecoflex 00-10 as Layer 1 and Dragon Skin 10 Medium as Layer 2, the hierarchical architectural design results in significant attenuation in the vibration amplitude. The amplitude at the edge of the haptic unit can be attenuated to 8.86\% and 4\% in the embedded multi-layer design and the encapsulating multi-layer design. Similarly, adopting Ecoflex 00-10 as Layer 1 and Dragon Skin 30 as Layer 2 attenuates the amplitude sharply, compared with the single-layer structure of Dragon Skin 30. For the embedded multi-layer design and the encapsulating multi-layer design, the optimized design decreases the amplitude at the edge of the haptic unit to 15.6\% and 5.17\%. These results indicate that the hierarchical architectural design has a significant effect on amplitude modulation. By changing the original single-layer structure to a two-layer structure, a significant reduction in vibration amplitude can be achieved.

\subsection{Ablation Study}
\label{exp:ab}
For the two types of hierarchical architectural design proposed in this study, using different materials for Layer 1 and Layer 2 results in varying amplitude modulation effects.

For both hierarchical architectural designs, when Ecoflex 00-10 is used as the material for Layer 1 and either Dragon Skin 10 Medium or Dragon Skin 30 for Layer 2, the amplitude of vibration waves passing through the layers is attenuated compared to the control group. Additionally, for encapsulating multi-layer design, when Dragon Skin 10 Medium serves as Layer 1 and Dragon Skin 30 as Layer 2, a similar modulation effect on the amplitude is observed. These observations suggest that when the elastic modulus of Layer 1 is lower than that of Layer 2, there is an attenuation effect on the vibration amplitude. This is consistent with Constraint \ref{eq:constraints} of the theoretical model.

On the contrary, when the elastic modulus of Layer 1 is higher than that of Layer 2, there is an amplification effect on the vibration. For the encapsulating multi-layer design, when Dragon Skin 10 Medium is used as the material for Layer 1 and Ecoflex 00-10 for Layer 2, the amplitude of the vibration after passing through the layers is larger than that of the control group. Similarly, for the embedded multi-layer design, using either Dragon Skin 10 Medium or Dragon Skin 30 as Layer 1 and Ecoflex 00-10 as Layer 2 results in an amplification of the amplitude compared to the control group. When Dragon Skin 10 Medium is used as Layer 2, while Dragon Skin 30 is material for Layer 1, the amplitude increases to 187.7\% of the control group. This observation further validates Constraint \ref{eq:constraints} of the theoretical model from another perspective.

For hierarchical architectural designs, the greater the difference in elastic modulus between the two layers, the more effective the amplitude modulation effect becomes. In the encapsulating multi-layer design, the configuration using Dragon Skin 10 Medium as Layer 1 and Dragon Skin 30 as Layer 2 demonstrates lower amplitude reduction compared with the setup where Ecoflex 00-10 is Layer 1 and Dragon Skin 10 Medium is Layer 2. Additionally, when Layer 2 is consistently Dragon Skin 30, the setup with Ecoflex 00-10 as Layer 1 outperforms the setup with Dragon Skin 10 Medium as Layer 1. For the embedded multi-layer design, When the material of Layer 1 is Dragon Skin 30, using Ecoflex 00-10 for Layer 2 provides a slightly better effect on amplitude amplification than using Dragon Skin 10 Medium for Layer 2. These observations indicate that a greater disparity in the elastic moduli of the two layers leads to more effective modulation of amplitude, which is consistent with Constraint \ref{eq:constraints} of the theoretical model.

Figure \ref{fig:amplitude-i} and \ref{fig:amplitude-j} compare the effectiveness of the embedded multi-layer design and encapsulating multi-layer design. When Ecoflex 00-10 is applied as Layer 1 and Dragon Skin 10 Medium as Layer 2, the normalized amplitude of the embedded multi-layer design is 8.86\% at the edge of the unit, while the rate for encapsulating multi-layer design is 4\%. In configurations where Ecoflex 00-10 serves as Layer 1 and Dragon Skin 30 as Layer 2, the embedded multi-layer design and encapsulating multi-layer design reduces the amplitude to 15.67\% and 4.58\% of the control group, respectively. When Dragon Skin 10 Medium is used as Layer 1 and Ecoflex 00-10 as Layer 2, the corresponding normalized amplitudes at the edge of the unit are 60.9\% and 67.1\%, respectively. These findings suggest that the encapsulating multi-layer design shows a marginally better effect in amplitude modulation. 


By adopting different materials and design schemes, it is possible to amplify or reduce the amplitude of vibrations induced by motors. Arranging the haptic units in an array can provide stronger tactile sensations over a larger area, or a more compact structure with less mechanical crosstalk. Assembling units made of different designs together can change the perceived intensity weight of different haptic units, making the vibration sensation stronger in some areas and less noticeable in others.

\section{Conclusion}
\label{sec:con}
This paper proposed two types of hierarchical architectural design of a haptic unit aimed at modulating the vibration amplitude stimulated by an ERM motor. A modified elastic model was established as the basis for the optimized design. An experimental platform was built to evaluate the performance of the design and validate the correctness of the theoretical model. Future work includes arranging these tactile units to form an array-based vibrotactile interface and demonstrating the effectiveness of the interface in suppressing mechanical crosstalk.

\bibliographystyle{IEEEtran}
\bibliography{root}
\end{document}

%% file: sections/package.tex
\usepackage{array}
\usepackage{textcomp}
\usepackage{upgreek}
\usepackage{stfloats}
\usepackage{url}
\usepackage{verbatim}
\usepackage[space]{cite}
\usepackage[english]{babel}

\makeatletter
\let\NAT@parse\undefined
\makeatother
\usepackage{hyperref}
\hypersetup{
    colorlinks=true,
    linkcolor=red,
    filecolor=magenta,      
    urlcolor=blue,
    citecolor=blue,
}
\usepackage{graphicx}
\usepackage{color}
\usepackage{tabularx} 
\usepackage{threeparttablex} 
\usepackage{multirow}
\usepackage{multirow}
\usepackage{makecell}
\usepackage[table,xcdraw]{xcolor}
\usepackage[normalem]{ulem}
\useunder{\uline}{\ul}{}

\usepackage{amsmath,amsfonts}
\usepackage{algorithmic}
\usepackage[ruled,vlined,linesnumbered]{algorithm2e}
\usepackage{amssymb}
\usepackage[normalem]{ulem}
\usepackage{CJKutf8}

\usepackage{subcaption}
